\documentclass[10pt,twocolumn,letterpaper]{article}

\usepackage[pagenumbers]{iccv} 

\usepackage{arydshln} 

\usepackage{graphicx}
\usepackage{amsmath}
\usepackage{amssymb}
\usepackage{dashrule}
\usepackage{booktabs}
\usepackage{array, multirow}
\usepackage{mathtools}
\usepackage{enumitem}
\usepackage{float}
\usepackage{makecell}
\usepackage[normalem]{ulem}
\useunder{\uline}{\ul}{}
\usepackage{bbm}

\usepackage{pifont}
\newcommand{\cmark}{\ding{51}}
\newcommand{\xmark}{\ding{55}}

\usepackage[dvipsnames]{xcolor, colortbl}

\newcommand{\gc}{\color{gray!80}}

\newcommand{\gr}[1]{\textcolor{gray}{#1}}
\definecolor{lavendar}{HTML}{F6DDF3}
\definecolor{lightGray}{HTML}{ECECEC}
\definecolor{needleGreen}{HTML}{B6D9C5}
\definecolor{longBlue}{HTML}{BBD9EC}
\definecolor{shortPink}{HTML}{FBCDCF}
\definecolor{avgColor}{HTML}{E7C97F}
\usepackage{lipsum}

\definecolor{iccvblue}{rgb}{0.21,0.49,0.74}
\usepackage[pagebackref,breaklinks,colorlinks,allcolors=iccvblue]{hyperref}

\def\paperID{4873} 
\def\confName{ICCV}
\def\confYear{2025}

\title{Multi-Granular Spatio-Temporal Token Merging for \\ Training-Free Acceleration of Video LLMs}

\author{
Jeongseok Hyun$^1$\footnotemark[1]{\qquad} Sukjun Hwang$^2${\qquad} Su Ho Han$^1${\qquad} Taeoh Kim$^3${\qquad} Inwoong Lee$^3${\qquad}
\vspace{0mm}\\ Dongyoon Wee$^3${\qquad} Joon-Young Lee$^4${\qquad} Seon Joo Kim$^1$\footnotemark[2]{\qquad} Minho Shim$^3$\footnotemark[2]
\vspace{2mm}\\ $^1$Yonsei University\qquad $^2$Carnegie Mellon University\qquad $^3$NAVER Cloud\qquad $^4$Adobe Research
\vspace{-3mm}
}

\begin{document}
\maketitle
\footnotetext[1]{This work was done during an internship at NAVER Cloud.}
\footnotetext[2]{Co-corresponding authors.}

\begin{abstract}
Video large language models (LLMs) achieve strong video understanding by leveraging a large number of spatio-temporal tokens, but suffer from quadratic computational scaling with token count.
To address this, we propose a training-free spatio-temporal token merging method, named STTM.
Our key insight is to exploit local spatial and temporal redundancy in video data which has been overlooked in prior work.
STTM first transforms each frame into multi-granular spatial tokens using a coarse-to-fine search over a quadtree structure, then performs directed pairwise merging across the temporal dimension.
This decomposed merging approach outperforms existing token reduction methods across six video QA benchmarks.
Notably, STTM achieves a 2× speed-up with only a 0.5\% accuracy drop under a 50\% token budget, and a 3× speed-up with just a 2\% drop under a 30\% budget.
Moreover, STTM is query-agnostic, allowing KV cache reuse across different questions for the same video.
The project page is available at 
\href{https://www.jshyun.me/projects/sttm}{https://www.jshyun.me/projects/sttm}.
\end{abstract}
    
\section{Introduction}
\label{sec:intro}

\begin{figure}[t!]
\centering
\includegraphics[width=\linewidth]{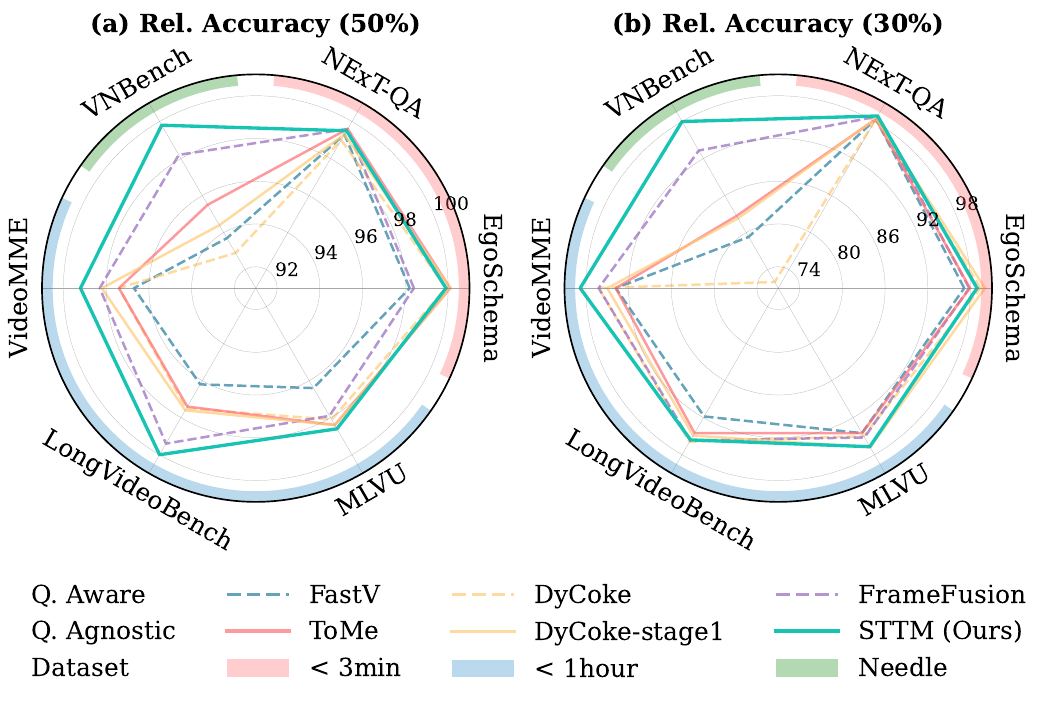}

\vspace{-1mm}

\includegraphics[width=\linewidth]{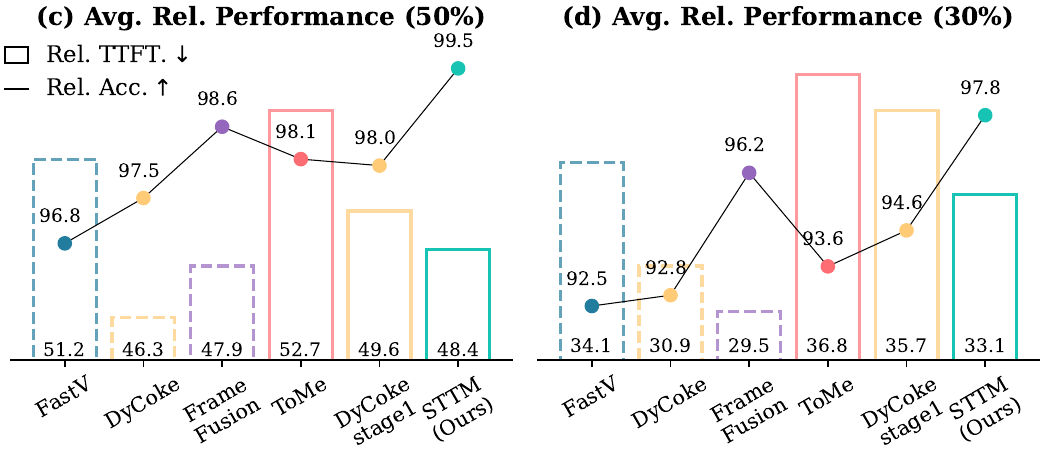}

\vspace{-3mm}

\caption{
Comparison of training-free token reduction methods using LLaVA-Video-7B under 50\% and 30\% pre-filling token budgets.
Query-aware (Q. Aware) methods require re-computation for each new query, whereas query-agnostic methods support KV-cache reuse.
The evaluated video QA datasets cover short ($<$3 min), long ($<$1 hr), and needle-in-a-haystack (NIAH) videos.
\textbf{(a, b)}: Per-dataset accuracy.
\textbf{(c, d)}: Average results across all.
}
\label{fig:teaser}

\vspace{-5mm}

\end{figure}

Integrating visual understanding capabilities into Large Language Models (LLMs) has led to significant advances in multimodal systems~\cite{touvron2023llama, dubey2024llama3, bai2023qwenvl, yang2024qwen2, wang2024qwen2vl, google2024gemini1_5, openai2024gpt4o}.
However, Video LLMs, which extend these capabilities to video understanding, face unique computational challenges due to the inherently large number of visual tokens required to represent spatio-temporal information~\cite{song2024moviechat, ren2024timechat, zhang2023videollama, maaz2023videochatgpt, li2023videochat, xue2025longvila}.

Early video LLMs~\cite{zhang2023videollama, li2023videochat, li2024mvbench, song2024moviechat} reduce token count by training abstractors (\eg, Q-Former~\cite{li2023blip2}) to compress visual information.
FlashAttention (FA)~\cite{dao2022flashattention1, dao2024flashattention2} reduces attention's quadratic memory cost to linear, enabling the training of long-context LLMs.
Building on this, LLaVA-style~\cite{liu2023llava} video LLMs use linear projectors that preserve high-resolution spatial features and deliver strong performance~\cite{zhang2024longva, zhang2024llavavideo178k}.
More recently, Ring Attention~\cite{liu2024ringattn} extends FA~\cite{dao2024flashattention2} across GPUs, enabling very long-context models, such as LWM~\cite{liu2024lwm} and LongVILA~\cite{xue2025longvila}, to process hour-long videos with high spatial and temporal resolution features.

Long latency remains a key bottleneck for deploying video LLMs.
Attention FLOPs still scale quadratically with token count, and long video contexts must be processed in full to pre-fill Key-Value (KV) states before answering any question~\cite{pope2023kvcache}.
In practice (\eg, Gemini~\cite{google_api_caching}), KV states for the video are cached to avoid recomputation for subsequent queries, 
supporting efficient multi-turn querying over a shared video context.
However, existing training-free token reduction methods~\cite{huang2024prunevid, chen2024fastv, fu2024framefusion, tao2025dycoke}, which avoid the overhead of retraining, overlook this scenario and instead focus on query-aware strategies that discard video tokens based on attention scores between the video and the query, without preserving KV cache reusability.

Designing a query-agnostic token reduction method is essential for enabling KV cache reuse, but it is challenging due to the lack of signals to guide token selection.
VideoMAE~\cite{tong2022videomae,feichtenhofer2022masked} demonstrates that videos can be reconstructed from a lower token ratio than images, highlighting the inherently redundant nature of video content due to its spatio-temporal continuity.
However, existing training-free token reduction methods do not explicitly leverage this spatio-temporal structure~\cite{chen2024fastv, fu2024framefusion, shang2024prumerge, wan2024lookm, tao2025dycoke}.
To address this, we propose a novel spatio-temporal token merging method, named \textit{STTM}, which is applied at an early layer of the LLM.
STTM performs decomposed merging: it first merges tokens along the spatial dimension, followed by merging along the temporal dimension.

To perform spatial merging, we build a multi-level quadtree for each video frame, linking each parent node to its four child nodes.
A parent node's token is retained at the coarsest level when its similarity to all four child nodes exceeds a threshold.
Regions containing fine-grained details -- indicated by low similarity -- are subdivided to the next finer level to preserve high-frequency information.
Consequently, each frame is represented by multi-granular spatial tokens, capturing both coarse and fine detail.

Extending spatial merging to the spatio-temporal domain is not trivial, due to varying spatial granularities across frames.
To handle this challenge, we exploit spatio-temporal locality by restricting merging candidates to spatially overlapping tokens between consecutive frames.
Unlike prior temporal-only methods that compare single-granularity tokens across adjacent frames~\cite{fu2024framefusion}, our approach fully leverages the spatial structure and temporal continuity in video data to guide token merging.

After comparing spatially overlapping tokens across time, similar token pairs are chained into spatio-temporal graphs.
Within each graph, nodes are merged into a single token representing a token tracklet, with merging directed toward the earlier frame.
This strategy accumulates temporal changes into the token where the content first appears.

The same region may be represented at different spatial levels across frames, leading to two cases: many-to-one (fine-to-coarse) and one-to-many (coarse-to-fine).
The many-to-one case is straightforward: merging fine tokens into a single coarse token in an early frame.
The one-to-many case is complicated.
Ideally, we would select the most similar fine-grained token as the destination, but this requires per-region comparisons that are difficult to vectorize and inefficient on GPUs.
To address this, we approximate the merge direction by selecting the top-left token among candidates.
This approximation allows us to implement temporal merging using a vectorized union-find algorithm~\cite{tarjan1975unionfind}, leading to efficient parallel processing.

Comprehensive experimental results across six video QA benchmarks demonstrate that our proposed method effectively reduces video tokens while maintaining high performance.
As shown in \cref{fig:teaser}, STTM outperforms both query-aware and query-agnostic methods overall.
In the average results \cref{fig:teaser}~(c, d), it achieves the highest accuracy with competitive latency under both 50\% and 30\% token budgets.
A closer look at the per-dataset accuracy in \cref{fig:teaser}~(a, b) further highlights its robustness, particularly on the challenging NIAH and long video datasets.
Moreover, we validate STTM's generalization by applying it to other LLMs and a large-scale 72B Video LLM.

\section{Related Work}
\label{sec:relworks}

\subsection{Spatio-Temporal Redundancy in Videos}
Videos exhibit strong spatio-temporal locality -- information within nearby regions in space and time is often redundant.
This property has long been exploited across various domains.
In deep learning, CNNs~\cite{lecun1989handwritten, he2016resnet} leverage spatial locality and are extended to 3D convolutions~\cite{tran2015c3d, carreira2017i3d} to model spatio-temporal correlations in video.
In transformers, MAE~\cite{he2022masked} and VideoMAE~\cite{tong2022videomae, feichtenhofer2022masked} show that reconstructing masked regions using local context is an effective self-supervised learning objective, implicitly benefiting from redundancy.

Similarity, video compression methods explicitly eliminate redundancy. Quadtree-based partitioning~\cite{finkel1974quad, samet1984quadtree} adapts block resolution based on regional complexity~\cite{sullivan1994efficient, wiegand2003overview, sullivan2012overview}.
Temporal redundancy is also reduced, for example through the \textit{Skip} mode in H.264/AVC~\cite{wiegand2003overview}, which omits re-encoding regions that remain unchanged across frames.
Inspired by these principles, we propose a multi-granular spatio-temporal token merging method that explicitly compresses redundant video tokens by exploiting local similarity in both space and time.

\begin{figure*}[t!]
\centering
\includegraphics[width=\linewidth]{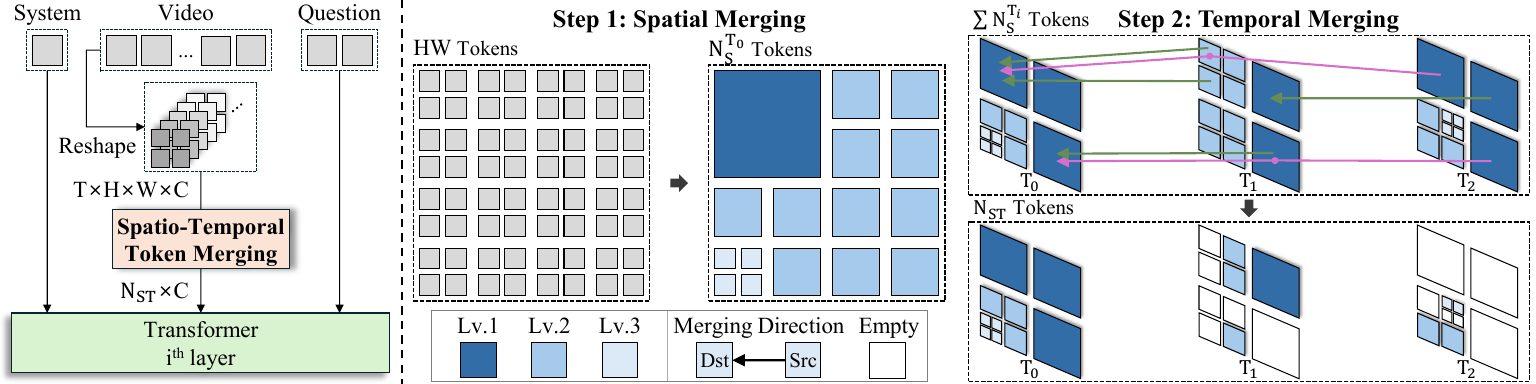}

\vspace{-3mm}

\caption{
\textbf{(Left)} Our spatio-temporal token merging method is a training-free, plug-and-play module that produces spatio-temporally multi-granular tokens.
\textbf{(Middle)} In step 1, tokens are merged based on spatial locality, where similar tokens within a 2D grid are combined into a single token.
\textbf{(Right)} In step 2, spatially multi-granular tokens are further merged along the temporal dimension, where similar tokens across frames are consolidated into their earliest occurrence.
The arrows indicate the direction of token merging.
The \textcolor[HTML]{678D54}{green lines} indicate merging over one timestep, and \textcolor[HTML]{D86FCC}{magenta lines} are merging over two timesteps.
The scale and the number of tokens are set for illustration.
}
\label{fig:overview}

\vspace{-5mm}

\end{figure*}

\subsection{Long Context Modeling in Video LLMs}

In the early stages, LLMs had a limited maximum context length; for instance, LLaMA~\cite{touvron2023llama} supported only 2048 tokens. To handle extensive video context, multimodal LLMs integrate abstractor architectures~\cite{cha2024honeybee, li2022blip, li2023blip2} that reduce visual tokens before passing them into the language model.
For video LLMs, Q-Former~\cite{li2023blip2} based architecture has been widely adopted to reduce the number of visual tokens from a video~\cite{li2023videochat, maaz2023videochatgpt, lin2023videollava, zhang2023videollama, song2024moviechat, li2024mvbench, wang2024videollamb, ren2024timechat, ryoo2024blip3video}, instead of using linear projectors, which preserve the original token count.

However, with the introduction of FlashAttention~\cite{dao2022flashattention1, dao2024flashattention2}, which reduces the memory complexity of attention from quadratic to linear in sequence length, linear projector-based approaches have recently gained more attention and showed better performance than abstractor-based approaches.
Furthermore, recent methods such as RingAttention~\cite{liu2024ringattn} have enabled video LLMs~\cite{liu2024lwm, xue2025longvila} to process up to one million video tokens. Following this trend, handling thousands of frames without downsampling is expected to become common in the future. However, such a large number of tokens significantly increases latency, highlighting the need for effective video token compression.

\subsection{Training-Free Token Reduction in Video LLMs}
Token reduction aims to identify and reduce highly redundant or less informative tokens, and has been researched in the architecture of Vision Transformer~\cite{bolya2023tome, choi2024vidtldr}.
Due to high computational costs in long video LLMs~\cite{xue2025longvila}, studying token reduction in the regime of multimodal LLMs is being spotlighted~\cite{shang2024prumerge, chen2024fastv, xing2024pyramiddrop, wan2024lookm, fu2024framefusion, tao2025dycoke, huang2024prunevid, jin2024chatunivi}.

In the initial works~\cite{chen2024fastv, shang2024prumerge}, visual tokens are reduced without considering the visual structure such as the 2D locality.
One-dimensional token reduction methods treating visual tokens as a sequential stream, similar to text tokens, without considering spatial or temporal relationships. FastV~\cite{chen2024fastv} reduces low-rank visual tokens based on attention patterns in the early layers. Additionally, it employs text queries to search attention weights, requiring a token reduction to be performed again for each new query. In contrast, our method is query-agnostic, allowing KV cache reuse in LLMs, making it more efficient for conversational and multi-question scenarios.

Temporal token reduction methods exploit the redundancy between video frames. FrameFusion~\cite{fu2024framefusion} demonstrates that the token similarity distribution condenses in deeper layers while preserving ranking consistency. 
It merges similar tokens between frames using similarity metrics and prunes unnecessary tokens based on attention scores. 
We argue that leveraging spatio-temporal characteristics, rather than operating in a single dimension, enables more effective and efficient token reduction.

\section{Methodology}
\label{sec:methodology}

We propose \textbf{S}patio-\textbf{T}emporal \textbf{T}oken \textbf{M}erging (STTM) that merges video tokens along spatial and temporal dimensions, producing multi-granular video tokens.
This module is training-free and can be easily plugged into an intermediate layer of an LLM.
As the STTM operates along the spatio-temporal dimension and outputs multi-granular tokens, it is designed as a single-pass operation, inserted into a single early layer of the transformer. 
Furthermore, our token reduction method is question-agnostic, enabling the reuse of KV caches across different questions for the same video.

\subsection{Overview of STTM Module}
As illustrated in \cref{fig:overview}, we use the video tokens from the previous layer's output to perform token merging.
After the merging process, the video tokens, $Z_V \in \mathbb{R}^{T\times H\times W\times C}$, are reduced into spatio-temporally merged tokens, $Z_{ST} \in \mathbb{R}^{N_{ST}\times C}$, where N$_{ST} \ll T\times H\times W$.

To exploit spatial locality, we first merge tokens that are similar within the 2D grid of each frame.
As labeled by Lv.1, Lv.2, and Lv.3 in \cref{fig:overview}, we define the different scales of the 2D grid for merging unit.
Based on these multi-scale merging grids, a large redundant region can be represented by a single coarse-grained token, achieving a high token reduction ratio.
On the other hand, fine-grained tokens are used for representing the region with large variation.
This spatial merging process results in spatially multi-granular tokens, $Z_{S}^{T_i} \in \mathbb{R}^{N_{S}^{T_i}\times C}$, where N$_{S}^{T_i} \ll H\times W$.

In videos, locality further extends to the spatio-temporal dimension.
Thus, we merge tokens by comparing their similarities at the same 2D region across consecutive frames.
When tokens remain similar over time, we chain them along a connected path and merge them into the earliest occurrence.
Tokens with different spatial granularities can also be connected across frames when they overlap. 
As depicted in step 2 of \cref{fig:overview}, some fine-grained tokens at $T_1$ are merged into a coarser token at $T_0$, while others remain separate due to low similarity.
Our spatial and temporal hybrid merging allows multi-granular tokens in both dimensions.

\begin{figure}[t!]
\centering
\includegraphics[width=\linewidth]{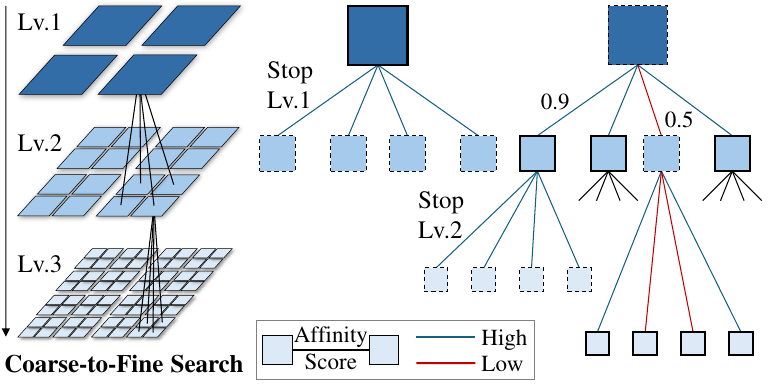}

\vspace{-2mm}

\caption{
A \textbf{coarse-to-fine spatial search} is performed using a quadtree structure.
If all four fine child nodes exhibit high similarity with the coarse parent node, the search process terminates, and the parent node is used to represent the corresponding region.
Otherwise, the search continues until the finest level is reached.
Here, the scale for each level is an example for illustration. 
}
\label{fig:spatial}

\vspace{-5mm}

\end{figure}

\subsection{Spatial Token Merging}

As illustrated in \cref{fig:spatial}, our approach employs a coarse-to-fine hierarchical search based on a quadtree data structure~\cite{finkel1974quad, samet1984quadtree}.
At every level, each region requires only four comparisons for the \textit{granularity decision process}; this process determines whether features at the current level sufficiently represent the region. If not, finer-grained features from the next level are adopted to represent the region without losing details.
This hierarchical method effectively balances computational efficiency and representation flexibility.
The quadtree structure organizes spatial tokens into a hierarchical representation, where each coarse-level parent node corresponds to four (2$\times$2) finer-level child nodes within its respective 2D grid region.
This hierarchical organization enables a structured approach to spatial token merging, preserving 2D locality, which is essential for subsequent temporal merging.

Building the quadtree follows three steps.
First, we initialize leaf nodes of the quadtree based on the initial token feature map, $Z^{T_i}_{lv} \in \mathbb{R}^{H \times W \times C}$.
Second, we construct a multi-scale representation by recursively downsampling the feature map until the feature map size reaches 2$\times$2, which we denote as Lv.1, representing the coarsest spatial resolution.
At each level, the feature map is downsampled by averaging spatially neighboring 2$\times$2 feature patches, yielding a coarser-level representation, $Z^{T_i}_{lv-1} \in \mathbb{R}^{\frac{H}{2} \times \frac{W}{2} \times C}$.
During this downsampling operation, we connect four leaf nodes from $Z^{T_i}_{lv}$ to a single parent node from $Z^{T_i}_{lv-1}$.
Third, using this precomputed densely connected quadtree, we perform a quadtree search as described in \cref{fig:spatial} and keep only the necessary nodes to represent each frame.

For each level, we compute the cosine similarity between the current level's nodes and their child nodes.
When the similarities of four child nodes are higher than the spatial threshold value, $\tau_S$, we can regard this region has low spatial detail and terminate the further searching process for the corresponding node.
As a result, all child nodes for this node are pruned in the quadtree.
In contrast, when any of the child nodes show low similarity, we prune the current node and instead use the child nodes to represent the corresponding region.
Thus, this quadtree search process can be regarded as the \textit{granularity decision process}.


The computational complexity of our quadtree-based spatial merging algorithm follows $\mathcal{O}(HW)$ per frame.
The quadtree search is an iterative process that performs four comparisons for each node at a given level. 
In the worst case scenario, where all nodes are subdivided down to the deepest level, the computational cost of spatial token merging, $C_{STM}$, is the geometric series following:
{
\setlength{\abovedisplayskip}{1mm}
\setlength{\belowdisplayskip}{1mm}
\begin{equation}
    C_{STM} = \sum_{i=1}^{\text{Lv}} \frac{HW}{4^i} \times 4 
\end{equation}
}
Asymptotically, $C_{STM}$ is bounded by the sum of infinite geometric series:
{
\setlength{\abovedisplayskip}{1mm}
\setlength{\belowdisplayskip}{1mm}
\begin{equation}
    \lim_{Lv \to \infty} C_{STM} = \frac{HW}{1-\frac{1}{4}} = \frac{4 H W}{3}
\end{equation}
}
Notably, we implement the quadtree search algorithm in parallel at each level, ensuring that the actual running time remains linear in complexity.

\begin{table*}[t!]
\aboverulesep=0ex
\belowrulesep=0ex
\setlength{\tabcolsep}{1pt} 
\centering
\resizebox{\linewidth}{!}{
\begin{tabular}{c | l | c | ccc | ccc | ccc | ccc | ccc | ccc | ccc}
\toprule
    \multirow{2}{*}{\begin{tabular}[c]{@{}l@{}} Token \\ Budget \end{tabular}} & \multirow{2}{*}{Method} & \multirow{2}{*}{\begin{tabular}[c]{@{}c@{}} Q. \\ Agn. \end{tabular}}
& \multicolumn{3}{c|}{\cellcolor{needleGreen} VNBench} & \multicolumn{3}{c|}{\cellcolor{longBlue} VideoMME} & \multicolumn{3}{c|}{\cellcolor{longBlue} LongVideoBench} & \multicolumn{3}{c|}{\cellcolor{longBlue} MLVU} &
\multicolumn{3}{c|}{\cellcolor{shortPink} EgoSchema} & \multicolumn{3}{c|}{\cellcolor{shortPink} NExT-QA} & \multicolumn{3}{c}{\cellcolor{avgColor} \textbf{\textit{Avg.}}}
\\
\cmidrule{4-24}
& & & Acc $\uparrow$ & TTFT $\downarrow$ & N$_\text{V}$ $\downarrow$ & Acc $\uparrow$ & TTFT $\downarrow$ & N$_\text{V}$ $\downarrow$ & Acc $\uparrow$ & TTFT $\downarrow$ & N$_\text{V}$ $\downarrow$ & Acc $\uparrow$ & TTFT $\downarrow$ & N$_\text{V}$ $\downarrow$ & Acc $\uparrow$ & TTFT $\downarrow$ & N$_\text{V}$ $\downarrow$ & Acc $\uparrow$ & TTFT $\downarrow$ & N$_\text{V}$ $\downarrow$ & Acc $\uparrow$ & TTFT $\downarrow$ & N$_\text{V}$ $\downarrow$ 
\\
\midrule
\midrule
\rowcolor{lightGray}
100\% & \textit{LLaVA-Video 7B} & \cmark &
        77.6 & 0.962 & 11149 &
        63.1 & 2.039 & 22086 &
        59.6 & 1.805 & 19624 &
        70.9 & 2.343 & 25088 &
        58.7 & 2.312 & 25069 &
        82.9 & 0.659 & 8116 &
        n.a & n.a & n.a
\\
\midrule
\multirow{6}{*}{\begin{tabular}[c]{@{}l@{}} 50\% \end{tabular}}
& \gr{+ FastV} &  &
    \gr{93.7} & \gr{52.3} & \gr{50.0} &
    \gr{96.7} & \gr{50.7} & \gr{50.0} &
    \gr{96.2} & \gr{50.8} & \gr{50.0} &
    \gr{96.4} & \gr{49.7} & \gr{50.0} &
    \gr{98.2} & \gr{50.4} & \gr{50.0} &
    \gr{99.3} & \gr{53.5} & \gr{50.0} &
    \gr{96.8} & \gr{51.2} & \gr{50.0}
\\
& \gr{+ DyCoke} &  & 
    \gr{92.9} & \gr{47.6} & \gr{49.4} & 
    \gr{97.4} & \gr{44.7} & \gr{47.9} & 
    \gr{97.5} & \gr{45.4} & \gr{48.5} & 
    \gr{98.1} & \gr{44.7} & \gr{47.7} & 
    \gr{99.9} & \gr{45.4} & \gr{47.7} & 
    \gr{99.0} & \gr{50.0} & \gr{49.3} & 
    \gr{97.5} & \gr{46.3} & \gr{48.4}
\\
& \gr{+ FrameFusion} &  &
    \gr{98.2} & \gr{49.0} & \gr{49.4} &
    \gr{98.3} & \gr{47.1} & \gr{48.7} &
    \gr{99.4} & \gr{47.2} & \gr{49.1} &
    \gr{97.9} & \gr{46.3} & \gr{48.4} &
    \gr{98.4} & \gr{47.0} & \gr{49.1} &
    \gr{99.6} & \gr{50.9} & \gr{49.6} &
    \gr{98.6} & \gr{47.9} & \gr{49.0}
\\ \cmidrule{2-24}
& + ToMe & \cmark &
    95.5 & 53.8 & 50.0 & 
    97.4 & 51.1 & 50.0 & 
    97.4 & 52.6 & 50.0 & 
    98.4 & 50.9 & 50.0 &
    100.1 & 51.9 & 50.0 &
    99.6 & 56.1 & 50.0 &
    98.1 & 52.7 & 50.0
\\
& + DyCoke-stage1 & \cmark & 
    94.4 & 51.4 & 49.4 & 
    98.2 & 48.1 & 47.9 & 
    97.6 & 48.6 & 48.5 & 
    98.4 & 47.9 & 47.7 & 
    100.1 & 48.4 & 47.7 & 
    99.3 & 53.0 & 49.3 &
    98.0 & 49.6 & 48.4
\\
\rowcolor{lavendar}\cellcolor{white}
& + STTM (Ours) & \cmark & 
    99.8 & 47.3 & 45.5 &
    99.2 & 50.1 & 47.1 &
    100.0 & 49.6 & 45.6 &
    98.6 & 49.2 & 48.6 &
    99.9 & 45.2 & 42.8 &
    99.5 & 48.9 & 43.7 &
    \textbf{99.5} & \textbf{48.4} & \textbf{45.5}
\\
\midrule
\multirow{6}{*}{\begin{tabular}[c]{@{}l@{}} 30\% \end{tabular}}
& \gr{+ FastV} &  & 
    \gr{79.3} & \gr{34.9} & \gr{30.0} & 
    \gr{93.8} & \gr{33.5} & \gr{30.0} & 
    \gr{91.8} & \gr{33.8} & \gr{30.0} & 
    \gr{94.5} & \gr{32.7} & \gr{30.0} & 
    \gr{97.1} & \gr{33.4} & \gr{30.0} & 
    \gr{98.4} & \gr{36.4} & \gr{30.0} & 
    \gr{92.5} & \gr{34.1} & \gr{30.0}
\\
& \gr{+ DyCoke} &  & 
    \gr{72.0} & \gr{32.0} & \gr{33.0} & 
    \gr{96.2} & \gr{29.3} & \gr{31.2} & 
    \gr{95.9} & \gr{30.2} & \gr{32.1} & 
    \gr{95.0} & \gr{29.2} & \gr{31.1} & 
    \gr{99.4} & \gr{30.0} & \gr{31.1} & 
    \gr{98.3} & \gr{34.7} & \gr{33.2} & 
    \gr{92.8} & \gr{30.9} & \gr{31.9}
\\
& \gr{+ FrameFusion} &  & 
    \gr{93.3} & \gr{30.3} & \gr{28.1} & 
    \gr{96.2} & \gr{28.5} & \gr{27.4} & 
    \gr{95.7} & \gr{28.8} & \gr{27.9} & 
    \gr{95.2} & \gr{27.8} & \gr{27.0} & 
    \gr{97.8} & \gr{28.4} & \gr{27.4} & 
    \gr{98.8} & \gr{33.0} & \gr{28.4} & 
    \gr{96.2} & \gr{29.5} & \gr{27.7}
\\ \cmidrule{2-24}
& + ToMe & \cmark &
    82.7 & 37.8 & 30.0 &
    93.8 & 35.3 & 30.0 &
    94.5 & 36.5 & 30.0 &
    94.5 & 35.0 & 30.0 &
    97.9 & 36.1 & 30.0 &
    98.4 & 40.2 & 30.0 &
    93.6 & 36.8 & 30.0
\\
& + DyCoke-stage1 & \cmark & 
    82.3 & 37.2 & 33.3 &
    95.0 & 34.3 & 31.2 &  
    94.9 & 35.0 & 32.1 &
    96.8 & 34.0 & 31.1 &
    100.0 & 34.6 & 31.1 & 
    98.5 & 38.9 & 33.2 &
    94.6 & 35.7 & 32.0
\\
\rowcolor{lavendar}\cellcolor{white}
& + STTM (Ours) & \cmark & 
    98.0 & 31.1 & 25.8 &
    98.8 & 31.4 & 25.8 &
    95.6 & 34.1 & 28.3 &
    96.7 & 32.8 & 29.2 &
    98.9 & 33.4 & 29.1 &
    98.9 & 35.7 & 27.3 &
    \textbf{97.8} & \textbf{33.1} & \textbf{27.6}
\\
\bottomrule
\end{tabular}
}
\vspace{-3mm}
\caption{
Comparison of training-free token reduction methods using LLaVA-Video-7B under 50\% and 30\% pre-filling token budgets.
Token-reduced results are reported relative to the \colorbox{lightGray}{result with 100\%}.
}
\label{tab:main_llava_video_7b}
\vspace{-3mm}
\end{table*}

\subsection{Temporal Token Merging}

To capture the spatio-temporal redundancy, we compute the similarity of tokens representing the same region in neighboring two frames and make the connection between pairs of tokens exhibiting similarity higher than $\tau_T$.
As we merge into the earlier tokens, this process results in directed graphs, where the source node is from $T+1$, and the destination node is at $T$.
Based on these directed graphs built from consecutive frames, we obtain connected graphs and identify the common root node for all nodes in the connected graph.
The common root node and other nodes are set as destination and source, respectively, for merging. 
So, the nodes can be merged not only in consecutive two-frame sequences but also in the far distance.
As shown in \cref{fig:overview} (right), a token representing the top-left token at $T_2$ is merged into the token at $T_0$, while the top-right token at $T_2$ is merged into the token at $T_1$.
This merging process results in spatio-temporally multi-scale tokens.

Since spatial merging results in varying scales of tokens, we cannot directly merge the tokens at the same exact spatial location across frames.
Instead, we merge the tokens under their spatio-temporally overlapping regions.
For example, in \cref{fig:overview} step 2, top-left region is expressed in varying scales across frames.
We compare the similarity between a single Lv.1 token at $T_0$ and $2\times 2$ Lv.2 tokens at $T_1$. 
Two Lv.2 tokens at $T_1$ with high similarity are merged into the token at $T_0$, while the other two remain to represent a change in $T_1$'s top-left region.

During temporal merging, we set the direction of merging as tokens at the earlier frame.
In the case of top-left at $T_1$ and $T_2$, the token at $T_2$ is coarser than $T_1$ and has four possible destination paths for merging.
When there are multiple connections with similarity higher than $\tau_T$, we provide the priority to the token located towards top-left.
We simplify the choice of choosing the top-left merging token for such multiple destination cases primarily for parallel indexing implementation, although merging with the most similar token is the naive method.

The computational complexity of our temporal merging algorithm follows $\mathcal{O}(THW)$ per video.
Suppose the worst case is that spatial tokens of all frames are in the finest level, incurring $(T-1) \times H\times W$ comparisons.
However, we can expect more efficiency gain since spatial merging results in fewer tokens at every frame, reducing the comparisons.
Computing the common root node can also be achieved in $\mathcal{O}(THW)$ by adapting the union-find algorithm~\cite{tarjan1975unionfind} into a vectorized form.

\subsection{Token Reordering after Merging}
\label{subsec:ordering}

After spatio-temporal token merging, the resulting video tokens form a graph structure.
To serve as input for an LLM, this graph must be linearized into a one-dimensional sequence.
In our reordering strategy, we prioritize tokens based on two criteria.
First, spatial ordering follows a Z-shaped scan based on each token's top-left coordinate.
Second, temporal ordering gives precedence to tokens from earlier frames over those from later ones.

Once reordered into a 1D sequence, handling positional embeddings becomes crucial.
Recent LLMs typically use rotary positional embeddings (RoPE)~\cite{su2024rope}, which apply a rotation matrix computed from each token's position index.
We evaluate three strategies for assigning RoPE after merging:
(1) Merged RoPE, which averages the RoPEs of merged tokens;
(2) Survived RoPE, which retains the original RoPEs of the surviving tokens;
(3) Reassigned RoPE, which reassigns position IDs based on the new token order for alignment with the original positional encoding.


\begin{table*}[t!]
\aboverulesep=0ex
\belowrulesep=0ex
\setlength{\tabcolsep}{1pt} 
\centering
\resizebox{\linewidth}{!}{
\begin{tabular}{c | l | c | ccc | ccc | ccc | ccc | ccc | ccc | ccc}
\toprule
    \multirow{2}{*}{\begin{tabular}[c]{@{}l@{}} Token \\ Budget \end{tabular}} & \multirow{2}{*}{Method} & \multirow{2}{*}{\begin{tabular}[c]{@{}c@{}} Q. \\ Agn. \end{tabular}}
& \multicolumn{3}{c|}{\cellcolor{needleGreen} VNBench} & \multicolumn{3}{c|}{\cellcolor{longBlue} VideoMME} & \multicolumn{3}{c|}{\cellcolor{longBlue} LongVideoBench} & \multicolumn{3}{c|}{\cellcolor{longBlue} MLVU} &
\multicolumn{3}{c|}{\cellcolor{shortPink} EgoSchema} & \multicolumn{3}{c|}{\cellcolor{shortPink} NExT-QA} & \multicolumn{3}{c}{\cellcolor{avgColor} \textbf{\textit{Avg.}}}
\\
\cmidrule{4-24}
& & & Acc $\uparrow$ & TTFT $\downarrow$ & N$_\text{V}$ $\downarrow$ & Acc $\uparrow$ & TTFT $\downarrow$ & N$_\text{V}$ $\downarrow$ & Acc $\uparrow$ & TTFT $\downarrow$ & N$_\text{V}$ $\downarrow$ & Acc $\uparrow$ & TTFT $\downarrow$ & N$_\text{V}$ $\downarrow$ & Acc $\uparrow$ & TTFT $\downarrow$ & N$_\text{V}$ $\downarrow$ & Acc $\uparrow$ & TTFT $\downarrow$ & N$_\text{V}$ $\downarrow$ & Acc $\uparrow$ & TTFT $\downarrow$ & N$_\text{V}$ $\downarrow$ 
\\
\midrule
\midrule
\rowcolor{lightGray}
100\% & \textit{LLaVA-OV 7B} & \cmark &
68.8 & 0.922 & 11149 & 
59.0 & 1.904 & 22086 & 
56.3 & 1.712 & 19624 &
67.5 & 2.224 & 25088 &
61.3 & 2.216 & 25069 &
80.6 & 0.630 & 8116 &
n/a & n/a & n/a
\\
\midrule
\multirow{6}{*}{\begin{tabular}[c]{@{}l@{}} 50\% \end{tabular}}
& \gr{+ FastV} &  &
\gr{95.3} & \gr{50.7} & \gr{50.0} &
\gr{99.0} & \gr{49.1} & \gr{50.0} &
\gr{99.7} & \gr{49.2} & \gr{50.0} &
\gr{99.9} & \gr{48.1} & \gr{50.0} &
\gr{99.4} & \gr{48.8} & \gr{50.0} &
\gr{99.5} & \gr{52.3} & \gr{50.0} &
\gr{98.8} & \gr{49.7} & \gr{50.0}
\\
& \gr{+ DyCoke} &  & 
\gr{94.3} & \gr{45.3} & \gr{49.4} &
\gr{101.8} & \gr{43.7} & \gr{47.9} &
\gr{102.8} & \gr{43.2} & \gr{48.5} &
\gr{101.8} & \gr{42.3} & \gr{47.7} &
\gr{100.8} & \gr{42.8} & \gr{47.7} &
\gr{99.7} & \gr{47.8} & \gr{49.3} &
\gr{100.2} & \gr{44.2} & \gr{48.4}
\\
& \gr{+ FrameFusion} &  &
\gr{98.6} & \gr{47.2} & \gr{49.8} &
\gr{101.4} & \gr{45.4} & \gr{49.4} &
\gr{100.5} & \gr{45.5} & \gr{49.6} &
\gr{101.0} & \gr{44.7} & \gr{49.3} &
\gr{99.1} & \gr{45.0} & \gr{49.9} &
\gr{100.2} & \gr{49.5} & \gr{50.0} &
\gr{100.1} & \gr{46.2} & \gr{49.7}
\\ \cmidrule{2-24}
& + ToMe & \cmark &
97.5 & 52.4 & 50.0 & 
101.4 & 50.6 & 50.0 &
102.5 & 50.5 & 50.0 & 
102.0 & 49.5 & 50.0 &
101.6 & 50.0 & 50.0 & 
99.4 & 54.4 & 50.0 &
100.7 & 51.2 & 50.5
\\
& + DyCoke-stage1 & \cmark & 
94.6 & 49.7 & 49.4 &
100.6 & 47.9 & 47.9 &
103.3 & 48.0 & 48.5 &
102.5 & 46.7 & 47.7 &
100.8 & 47.3 & 47.7 &
100.0 & 51.8 & 49.3 &
100.3 & 48.6 & 48.4
\\
\rowcolor{lavendar}\cellcolor{white}
& + STTM (Ours) & \cmark & 
103.6 & 43.8 & 44.2 &
102.8 & 40.6 & 37.5 &
102.4 & 47.0 & 44.6 & 
103.3 & 45.9 & 46.6 & 
100.6 & 43.9 & 43.6 & 
99.8 & 49.5 & 46.4 & 
\textbf{102.1} & \textbf{45.1} & \textbf{43.8}
\\
\midrule
\multirow{6}{*}{\begin{tabular}[c]{@{}l@{}} 30\% \end{tabular}}
& \gr{+ FastV} &  & 
\gr{86.1} & \gr{32.8} & \gr{30.0} &
\gr{98.6} & \gr{31.4} & \gr{30.0} &
\gr{98.1} & \gr{31.7} & \gr{30.0} &
\gr{97.2} & \gr{30.6} & \gr{30.0} &
\gr{99.3} & \gr{31.3} & \gr{30.0} &
\gr{98.8} & \gr{34.6} & \gr{30.0} &
\gr{96.4} & \gr{32.1} & \gr{30.0}
\\
& \gr{+ DyCoke} &  & 
\gr{78.4} & \gr{30.3} & \gr{33.3} &
\gr{101.4} & \gr{28.2} & \gr{31.2} &
\gr{99.3} & \gr{28.5} & \gr{32.1} &
\gr{101.2} & \gr{26.9} & \gr{31.1} &
\gr{101.2} & \gr{27.4} & \gr{31.1} &
\gr{98.9} & \gr{33.0} & \gr{33.2} &
\gr{96.7} & \gr{29.1} & \gr{32.0}
\\
& \gr{+ FrameFusion} &  & 
\gr{96.0} & \gr{28.3} & \gr{29.0} &
\gr{100.1} & \gr{26.2} & \gr{28.0} &
\gr{100.4} & \gr{26.6} & \gr{28.5} &
\gr{98.8} & \gr{25.6} & \gr{27.6} &
\gr{98.3} & \gr{26.3} & \gr{28.9} &
\gr{99.1} & \gr{31.0} & \gr{29.5} &
\gr{98.8} & \gr{27.3} & \gr{28.6}
\\ \cmidrule{2-24}
& + ToMe & \cmark &
86.0 & 36.0 & 30.0 &
101.4 & 34.1 & 30.0 &
100.3 & 34.3 & 30.0 &
102.4 & 33.3 & 30.0 &
101.3 & 33.8 & 30.0 & 
98.6 & 38.1 & 30.0 & 
98.3 & 34.9 & 30.0
\\
& + DyCoke-stage1 & \cmark & 
84.9 & 35.1 & 33.3 &
102.3 & 33.2 & 31.2 &
99.7 & 33.7 & 32.1 &
100.7 & 32.5 & 31.1 &
101.5 & 33.0 & 31.1 &
98.8 & 37.4 & 33.2 & 
98.0 & 34.2 & 32.0
\\
\rowcolor{lavendar}\cellcolor{white}
& + STTM (Ours) & \cmark & 
102.3 & 30.1 & 27.4 & 
102.6 & 31.6 & 27.3 &
100.5 & 33.3 & 30.0 &
101.4 & 33.0 & 31.8 &
100.7 & 25.7 & 22.4 &
98.9 & 38.1 & 33.1 &
\textbf{101.1} & \textbf{32.0} & \textbf{28.7}
\\
\bottomrule
\end{tabular}
}
\vspace{-3mm}
\caption{
Comparison of training-free token reduction methods using LLaVA-OneVision-7B.
Relative to \colorbox{lightGray}{100\% result}.
}
\label{tab:main_llava_onevision_7b}
\vspace{-1mm}
\end{table*}
\begin{table*}[t!]
\aboverulesep=0ex
\belowrulesep=0ex
\setlength{\tabcolsep}{1pt} 
\centering
\begin{minipage}{0.70\textwidth}
\resizebox{\linewidth}{!}{
    \begin{tabular}{c | l | ccc | ccc | ccc | ccc }
    \toprule
        \multirow{2}{*}{\begin{tabular}[c]{@{}l@{}} Token \\ Budget \end{tabular}} & \multirow{2}{*}{Method} & \multicolumn{3}{c|}{\cellcolor{needleGreen} VNBench} & \multicolumn{3}{c|}{\cellcolor{longBlue} VideoMME} & \multicolumn{3}{c|}{\cellcolor{longBlue} LongVideoBench} & \multicolumn{3}{c}{\cellcolor{avgColor} \textbf{\textit{Avg.}}}
    \\
    \cmidrule{3-14}
    
    & & Acc $\uparrow$ & TTFT $\downarrow$ & N$_\text{V}$ $\downarrow$ & Acc $\uparrow$ & TTFT $\downarrow$ & N$_\text{V}$ $\downarrow$ & Acc $\uparrow$ & TTFT $\downarrow$ & N$_\text{V}$ $\downarrow$ & Acc $\uparrow$ & TTFT $\downarrow$ & N$_\text{V}$ $\downarrow$ 
    \\
    \midrule
    \midrule
    \rowcolor{lightGray}
    100\% 
     &  \textit{Qwen2VL 7B} &
        66.4 & 2.438 & 22025 &
        61.8 & 10.745 & 74982 &
        56.8 & 10.597 & 72109 &
        n.a & n.a & n.a
    \\
    \midrule
    \multirow{3}{*}{\begin{tabular}[c]{@{}l@{}} 50\% \end{tabular}}
    & + ToMe &
        95.5 & 46.3 & 50.0 &
        100.1 & 42.0 & 50.0 &
        101.4 & 41.9 & 50.0 &
        99.0 & 43.4 & 50.0
    \\
    & + DyCoke-stage1 &
        98.1 & 43.0 & 49.6 &
        101.2 & 38.8 & 48.1 &
        101.2 & 39.1 & 48.8 &
        100.2 & 40.3 & \textbf{48.8}
    \\
    \rowcolor{lavendar}\cellcolor{white}
        & + STTM (Ours) & 
        105.2 & 29.8 & 48.9 &
        101.8 & 44.3 & 52.3 &
        101.1 & 43.5 & 51.5 &
        \textbf{102.7} & \textbf{39.2} & 50.9
    \\
    \midrule
    \multirow{3}{*}{\begin{tabular}[c]{@{}l@{}} 30\% \end{tabular}}
    & + ToMe &
    85.6 & 30.3 & 30.0 &
    99.3 & 26.4 & 30.0 &
    99.5 & 26.3 & 30.0 &
    94.8 & 27.7 & \textbf{30.0}
    \\
    & + DyCoke-stage1 & 
    81.6 & 29.0 & 33.4 &
    99.9 & 25.2 & 31.4 &
    101.2 & 25.4 & 32.3 & 
    94.2 & 26.5 & 32.4
    \\
    \rowcolor{lavendar}\cellcolor{white}
    & + STTM (Ours) & 
    100.4 & 17.2 & 30.6 &
    101.0 & 25.7 & 32.9 &
    100.1 & 23.3 & 27.7 &
    \textbf{100.5} & \textbf{22.1} & 30.4
    \\
    \bottomrule
    \end{tabular}
}
\vspace{-3mm}
\caption{Comparison using Qwen2VL-7B.
Relative to \colorbox{lightGray}{100\% result}.
}
\label{tab:main_qwen2vl_7b}
\vspace{-5mm}
\end{minipage}
\hfill
\begin{minipage}{0.28\linewidth}
\resizebox{1.0\linewidth}{!}{
    \begin{tabular}{c | l | ccc }
    \toprule
        \multirow{2}{*}{\begin{tabular}[c]{@{}l@{}} Token \\ Budget \end{tabular}} & \multirow{2}{*}{Method}
    & \multicolumn{3}{c}{\cellcolor{longBlue} VideoMME}
    \\
    \cmidrule{3-5}
    & & Acc $\uparrow$ & TTFT $\downarrow$ & N$_\text{V}$ 
    \\
    \midrule
    \midrule
    \rowcolor{lightGray}
    100\% & \textit{LLaVA-Video 72B} &
        70.5 & 17.698 & 22086
    \\
    \midrule
    \multirow{3}{*}{\begin{tabular}[c]{@{}l@{}} 50\% \end{tabular}}
    & + ToMe &
        100.1 & 47.6 & 50.0
    \\
    & + DyCoke-stage1 & 
        99.8 & 45.5 & 47.9
    \\
    \rowcolor{lavendar}\cellcolor{white}
    & + STTM (Ours) &
        \textbf{101.3} & \textbf{44.2} & \textbf{44.2}
    \\
    \midrule
    \multirow{3}{*}{\begin{tabular}[c]{@{}l@{}} 30\% \end{tabular}}
    & + ToMe & 
        97.1 & \textbf{29.3} & \textbf{30.0}
    \\
    & + DyCoke-stage1 & 
        98.3 & 30.2 & 31.2
    \\
    \rowcolor{lavendar}\cellcolor{white}
    & + STTM (Ours) & 
        \textbf{99.1} & 30.5 & 30.5
    \\
    \bottomrule
    \end{tabular}
}
\vspace{-3mm}
\caption{
Comparison using LLaVA-Video-72B.
Relative to \colorbox{lightGray}{100\% result}.
}
\label{tab:main_llava_video_72b}
\vspace{-5mm}
\end{minipage}
\end{table*}

\section{Experiments}

\subsection{Evaluation Setting}
\noindent \textbf{Datasets.}
To focus on visual understanding, we adopt an evaluation setting without subtitles. We evaluate methods on six diverse video QA benchmarks. These include short-form datasets, EgoSchema~\cite{mangalam2023egoschema} and NExT-QA~\cite{xiao2021nextqa}, as well as long-form datasets with hour-long videos: VideoMME~\cite{fu2025videomme}, LongVideoBench~\cite{wu2025longvideobench}, and MLVU~\cite{zhou2025mlvu}.
To rigorously evaluate fine-grained spatio-temporal understanding, we use VNBench~\cite{zhao2024vnbench}, a synthetic dataset simulating the \textit{needle in a haystack} task~\cite{kamradt2023needle}. It introduces subtle visual or textual ``needles'' into short segments of a video -- irrelevant to the original content but relevant to the question.

\noindent \textbf{Evaluation Metrics.}
We report standard accuracy for multiple-choice question answering.
For practicality, we include the average running time (in seconds) and the number of visual tokens (N$_V$).
Since our focus is on the pre-filling stage, we measure the time-to-first-token (TTFT).
In addition to absolute values, we provide relative values (R.) with respect to performance without token reduction. 

\noindent \textbf{Implementation Details.}
We sample video frames at 1 FPS, but uniformly sample frames if a video exceeds the maximum frame limit. To ensure coverage of injected needles in VNBench, we set the maximum number of frames to 180; for other datasets, we use a limit of 128 frames.
A single and four A100 80G GPUs are used for 7B and 72B models, respectively.
Our merging threshold values ($\tau_S$ and $\tau_T$) are empirically adjusted to approximately meet specific token budget for fair comparisons with other methods.

\subsection{Comparison with Existing Methods}
We compare other methods~\cite{chen2024fastv, fu2024framefusion, tao2025dycoke, bolya2023tome} under the same experimental setup.
ToMe~\cite{bolya2023tome} is applied within an LLM rather than its original ViT-based setting.
DyCoke-stage1 refers to a technique for pre-filling stage~\cite{tao2025dycoke}.
We report relative values here, with absolute values provided in Appendix.

\noindent \textbf{Performance of LLaVA-Video-7B.}
Our method outperforms both query-aware and query-agnostic methods on average across benchmarks (\cref{tab:main_llava_video_7b}).
On \colorbox{avgColor}{average}, it incurs only 0.5\% and 2.2\% relative accuracy drops under the 50\% and 30\% budgets, respectively.
On \colorbox{needleGreen}{VNBench} (30\% budget), other Q.Agn methods exhibit large drops (about 18\%), but it shows only a 2.0\% drop.
These results indicate that our method better preserves fine-grained spatio-temporal details.
All methods apply token reduction with $\mathcal{O}(N)$ complexity; thus, TTFT is dominated by attention in the LLM layers and scales with the number of retained tokens.
Since tokens are reduced at the ipnut or in the early LLM layers, all methods exhibit similar TTFT at each token budget.

\noindent \textbf{Generalization to Other MLLMs.}
We also evaluate our method with LLaVA-OneVision (\cref{tab:main_llava_onevision_7b}) and Qwen2VL (\cref{tab:main_qwen2vl_7b}), and it continues to outperform other methods on both MLLMs.
Notably, it even improves accuracy while using fewer tokens (under both 50\% and 30\% budgets).
We observe accuracy improvements of 1.1\% and 0.5\%, along with 3.1$\times$ and 4.5$\times$ speed-ups, for OneVision and Qwen2VL, respectively.
Qwen2VL consumes more visual tokens than LLaVA-based models, resulting in a greater latency reduction due to the quadratic complexity of attention.

\noindent \textbf{Scaling to 72B.}
Our method continues to outperform other methods on the 72B LLM under both 50\% and 30\% token budgets (\cref{tab:main_llava_video_72b}).
Notably, it even improves accuracy by 1.3\% while using only 44.2\% of the tokens.
The 72B model requires substantial processing time per request (\eg, an average of 17.7 seconds).
As a query-agnostic method, it performs token reduction without relying on the question, enabling reuse of the KV cache for the same video across different questions.
This improves deployment efficiency in multi-turn or multi-query scenarios.

\begin{table*}[t!]
\aboverulesep=0ex
\belowrulesep=0ex
\setlength{\tabcolsep}{3pt} 
\centering
\begin{minipage}[b]{0.40\textwidth}
\resizebox{\linewidth}{!}{
\begin{tabular}{c|cc|cc|cc}
\toprule
    \multirow{2}{*}{\begin{tabular}[c]{@{}c@{}} Token\\Granularity \end{tabular}} &
    \multirow{2}{*}{\begin{tabular}[c]{@{}c@{}} Lv.1\\Scale \end{tabular}} &
    \multirow{2}{*}{\begin{tabular}[c]{@{}c@{}} $\tau_S$ \end{tabular}} &
    \multicolumn{2}{c|}{\cellcolor{needleGreen} VNBench} & \multicolumn{2}{c}{\cellcolor{longBlue} VideoMME} \\
\cmidrule{4-7}
&&& R.Acc $\uparrow$ & R.N$_\text{V}$ $\downarrow$
& R.Acc $\uparrow$ & R.N$_\text{V}$ $\downarrow$
\\
\midrule
\midrule
Single & \multicolumn{2}{c|}{13$\times$13} & 82.1 & 86.2 & 99.2 & 86.2\\
\cdashline{1-7}
Multi & 2$\times$2 & 0.85 & 88.1 & 54.2 & 98.1 & 58.2 \\
Multi & 4$\times$4 & 0.85 & 99.9 & 82.0 & 99.5 & 83.0 \\
\hline
Single & \multicolumn{2}{c|}{11$\times$11} & 80.0 & 61.7 & 98.4 & 61.7 \\
\cdashline{1-7}
Multi & 2$\times$2 & 0.80 & 73.1 & 27.0 & 96.2 & 31.5 \\
Multi & 4$\times$4 & 0.80 & 98.7 & 57.2 & 99.2 & 59.9 \\
\hline
Single & \multicolumn{2}{c|}{9$\times$9} & 78.9 & 41.3 & 97.4 & 41.3 \\
\cdashline{1-7}
Multi & 2$\times$2 & 0.75 & 62.6 & 13.3 & 92.5 & 15.8 \\
Multi & 4$\times$4 & 0.75 & 93.0 & 35.8 & 97.5 & 38.9 \\
\hline
\multicolumn{3}{c|}{\gc FastV} & \gc 93.7 & \gc 50.0 & \gc 96.7 & \gc 50.0 \\
\bottomrule
\end{tabular}
}
\vspace{-3mm}
\caption{
Ablation study on the proposed multi-granular spatial token merging.
Lv.1: spatial scale of root nodes.
}
\label{tab:abl_root_node_res}
\vspace{-3mm}
\end{minipage}
\hfill
\begin{minipage}[b]{0.31\textwidth}
\resizebox{\linewidth}{!}{
\begin{tabular}{cc|cc|cc}
\toprule
   \multirow{2}{*}{\begin{tabular}[c]{@{}c@{}} STM \end{tabular}} & 
   \multirow{2}{*}{\begin{tabular}[c]{@{}c@{}} TTM \end{tabular}} &
   \multicolumn{2}{c|}{\cellcolor{needleGreen} VNBench} & \multicolumn{2}{c}{\cellcolor{longBlue} VideoMME} \\
\cmidrule{3-6}
&& R.Acc. $\uparrow$ & R.N$_\text{V}$ $\downarrow$
& R.Acc. $\uparrow$ & R.N$_\text{V}$ $\downarrow$
\\
\midrule
\midrule
\cmark & \xmark & 
    98.7 & 57.2 & 99.2 & 59.9
\\
\xmark & \cmark & 
    100.0 & 53.7 & 99.9 & 56.7
\\
\cmark & \cmark & 
    98.0 & 25.8 & 98.8 & 25.8
\\
\cdashline{1-6}
\multicolumn{2}{c|}{\cmark} & 
    73.3 & 27.0 & 95.5 & 32.0
\\
\bottomrule
\end{tabular}
}
\vspace{-3mm}
\caption{
Effectiveness of spatial and temporal merging modules.
Sequentially combining these two modules results in a synergistic effect.
Last row: joint merging.
}
\label{tab:abl_sm_and_tm}
\vspace{-2mm}
\end{minipage}
\hfill
\begin{minipage}[b]{0.26\textwidth}
\begin{minipage}[b]{\linewidth}
    \resizebox{\linewidth}{!}{
    \setlength{\tabcolsep}{5pt} 
    \begin{tabular}{l| ccc}
    \toprule
    Method & R.Acc $\uparrow$ & R.TTFT $\downarrow$ & R.N$_\text{V}$ $\downarrow$
    \\
    \midrule
    \midrule
    Optimal & \textbf{99.6} & 58.1 & 47.7
    \\
    Top-left & 99.2 & \textbf{50.1} & \textbf{47.1}
    \\
    \cdashline{1-4}
    Optimal & 97.7 & 39.6 & 27.7
    \\
    Top-left & \textbf{98.8} & \textbf{31.4} & \textbf{25.8}
    \\
    \bottomrule
    \end{tabular}
    }
    \vspace{-3mm}
    \caption{Temporal merging ablation}
    \label{tab:abl_merging}
\end{minipage}

\vspace{2mm}

\begin{minipage}[b]{\linewidth}
    \centering
    \resizebox{1.0\linewidth}{!}{
    \setlength{\tabcolsep}{5pt} 
    \begin{tabular}{c|c|c}
        \toprule
        \multirow{2}{*}{\begin{tabular}[c]{@{}c@{}} Positional \\ Embedding \end{tabular}}  & \cellcolor{needleGreen} VNB & \cellcolor{longBlue} VidMME  \\ \cmidrule{2-3}
        & R.Acc $\uparrow$ & R.Acc $\uparrow$ \\
        \midrule
        \midrule
            Merging & 96.0 & 95.7 \\
            Survival & 96.5 & 96.0 \\
            Reassignment & \textbf{98.0} & \textbf{98.8} \\
        \bottomrule
    \end{tabular}
    }
    \vspace{-3mm}
    \caption{Ablation study on positional embedding after merging.}
    \label{tab:abl_pos_emb}    
    \vspace{-3mm}
\end{minipage}
\end{minipage}
\end{table*}
\begin{table}[t!]
\aboverulesep=0ex
\belowrulesep=0ex
\setlength{\tabcolsep}{5pt} 
\centering
\resizebox{1.0\linewidth}{!}{
\begin{tabular}{c|ccc|ccc}
\toprule
   \multirow{2}{*}{\begin{tabular}[c]{@{}l@{}} LLM\\Layer \end{tabular}} &
   \multicolumn{3}{c|}{\cellcolor{needleGreen} VNBench} & \multicolumn{3}{c}{\cellcolor{longBlue} VideoMME} \\
\cmidrule{2-7}
& R.Acc $\uparrow$ & R.TTFT $\downarrow$ & R.N$_\text{V}$ $\downarrow$
& R.Acc $\uparrow$ & R.TTFT $\downarrow$ & R.N$_\text{V}$ $\downarrow$
\\
\midrule
\midrule
1 
    & 95.8 & \textbf{27.9} & 28.3
    & 96.0 & \textbf{26.5} & 26.4
\\
3 
    & 98.0 & 30.9 & 25.8
    & 98.8 & 31.3 & 25.8
\\
7 
    & 98.2 & 40.6 & \textbf{24.8}
    & 97.9 & 41.6 & 25.6
\\
19 
    & \textbf{99.5} & 75.5 & 31.7
    & \textbf{99.2} & 71.6 & \textbf{22.6}
\\
    \bottomrule
    \end{tabular}
}

\vspace{-3mm}

\caption{
Ablation study of token merging position.
}
\label{tab:abl_llm_loc}

\vspace{-3mm}
\end{table}

\begin{table}[t!]
\aboverulesep=0ex
\belowrulesep=0ex
\setlength{\tabcolsep}{3pt} 
\centering
\resizebox{1.0\linewidth}{!}{
\begin{tabular}{c|c|c|c|c|c|c | c}
\toprule
    G-Token & \cellcolor{needleGreen} VNB & \cellcolor{longBlue} VidMME & \cellcolor{longBlue} LVB & \cellcolor{longBlue} MLVU & \cellcolor{shortPink} EgoS & \cellcolor{shortPink} NExT & \cellcolor{lightGray} \textit{Avg.} ($\Delta$) \\
\midrule
\midrule
\cmark & 
    \textbf{78.2} & 62.6 & 58.9 &
    70.6 & 58.5 & 82.2 & 
    68.5
\\
\xmark & 
    77.6 & \textbf{63.1} & \textbf{59.6} &
    \textbf{70.9} & \textbf{58.7} & \textbf{82.9} &
    \textbf{68.8} ($+0.3$)
\\
\bottomrule
\end{tabular}
}

\vspace{-3mm}

\caption{
Effect of grid token removal on accuracy.
}
\label{tab:abl_grid_token}

\vspace{-5mm}
\end{table}

\subsection{Ablation Study}
We conduct ablation studies to validate effectiveness of the proposed components using LLaVA-Video-7B. 

\noindent \textbf{Multi-Granularity Spatial Tokens.}
To evaluate the effectiveness of our hierarchical spatial token merging method, we compare it against a single-granularity method using bilinear interpolation (\cref{tab:abl_root_node_res}).
On VideoMME, the single-granularity method performs competitively, even surpassing FastV~\cite{chen2024fastv} while using only 41.3\% of the tokens.
However, its accuracy drops significantly on VNBench which requires fine-grained spatial information.
In contrast, our quadtree-based multi-granularity merging maintains accuracy across both datasets.
At $\tau_S$ of 0.80, it incurs only 1.3\% and 0.8\% drop in relative accuracy with a token budget of approximately 50\%.
This highlights the advantage of incorporating multi-granularity spatial tokens for balancing compression efficiency and accuracy retention.

\noindent \textbf{Root Node Spatial Resolution.}
As shown in \cref{tab:abl_root_node_res}, using a coarser spatial scale (2$\times$2) for root node initialization leads to a lower relative N$_V$ at the same spatial threshold.
This indicates that more tokens are merged and each frame is represented by coarser spatial tokens.
However, this higher compression ratio comes at the cost of a significant accuracy drop compared to using root nodes with a 4$\times$4 spatial scale.
The effect is particularly pronounced at lower $\tau_S$, where the accuracy degradation is more severe, especially on VNBench.
This is due to the loss of spatial details, as quadtree subdivision from a 2$\times$2 root may terminate prematurely, producing tokens that are too coarse to retain essential visual information.
To address this, we adopt a 4$\times$4 spatial scale for root nodes, which enables lower $\tau_S$ for higher compression while preserving accuracy.

\noindent \textbf{Decomposed Spatio-Temporal Merging.}
In our algorithm, we first perform spatial merging, followed by temporal merging in sequence.
\cref{tab:abl_sm_and_tm} presents the impact of spatial and temporal merging, individually.
While both methods effectively reduce the number of tokens, applying them sequentially yields a synergistic effect, achieving a high token reduction ratio with only a marginal accuracy drop.
Additionally, we experiment with a joint spatio-temporal merging method based on an octree data structure, which partitions a long video into predefined cubic segments and recursively subdivides them at a 2$\times$2$\times$2 finer scale.
While this method shows promising results on VideoMME, it leads to a significant accuracy drop on VNBench, where rapid spatial changes occur within short time intervals. 
This suggests that rigid hierarchical partitioning across spatio-temporal dimensions is not effective in dynamic scenarios. 
Therefore, we adopt a decomposed merging strategy.

\noindent \textbf{Approximated Temporal Matching.}
As shown in \cref{tab:abl_merging}, selecting the most similar destination node does not consistently improve performance.
In contrast, using a top-left approximation enables a vectorized union-find algorithm~\cite{tarjan1975unionfind}, avoiding costly similarity checks, and leads to significant runtime improvement without drop in accuracy.

\noindent \textbf{Positional Embedding.}
\cref{tab:abl_pos_emb} shows the results of different strategies for handling positional embeddings during token merging within LLM layers.
We observe that preserving the original positional embeddings yields better performance than merging them.
Since Qwen2VL~\cite{wang2024qwen2vl} uses M-RoPE, which computes positions based on (t, y, x), we cannot apply the reassignment strategy and instead adopt the survival strategy.

\noindent \textbf{Merging Layer Position.}
We evaluate the impact of different LLM layer positions for video token merging (\cref{tab:abl_llm_loc}).
As observed in recent works~\cite{chen2024fastv, cai2024pyramidkv, xing2024pyramiddrop, yang2024pyramidinfer}, performing token merging at later layers yields higher accuracy but lower speed-up, due to increased full-token attention computations.
Merging tokens before the third LLM layer provides a good trade-off between accuracy and latency; we adopt this configuration for 7B models, while for the 72B model, we merge tokens before the first LLM layer.

\noindent \textbf{Grid Token.}
As shown in \cref{tab:abl_grid_token}, removing the special grid token used for video data in LLaVA-Video does not degrade accuracy across benchmarks.
To simplify the handling of video tokens during spatio-temporal merging, we omit such token in LLaVA-Video and OneVision.

\begin{figure}[t!]
\vspace{-2mm}
\centering
\includegraphics[width=\linewidth]{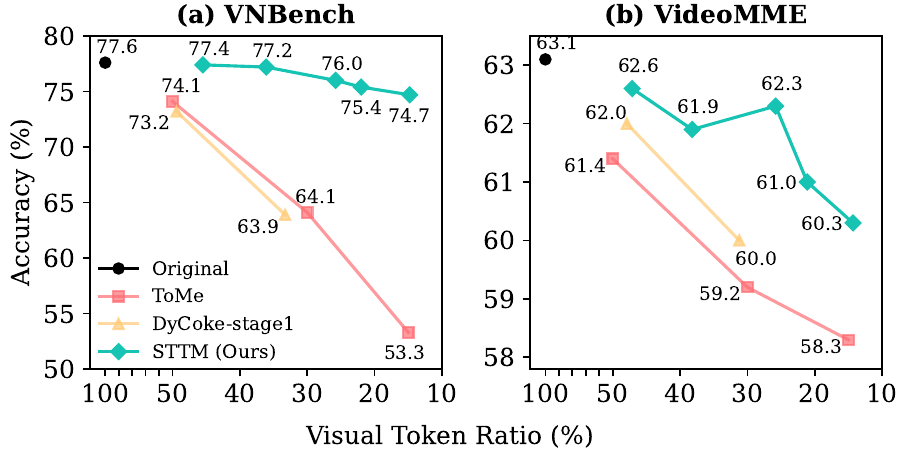}

\vspace{-1mm}

\caption{
Trade-off of accuracy and visual token retention ratio.
}
\label{fig:acc_token_tradeoff}

\vspace{-5mm}
\end{figure}

\begin{figure*}[t!]
\centering
\includegraphics[width=0.9\linewidth]{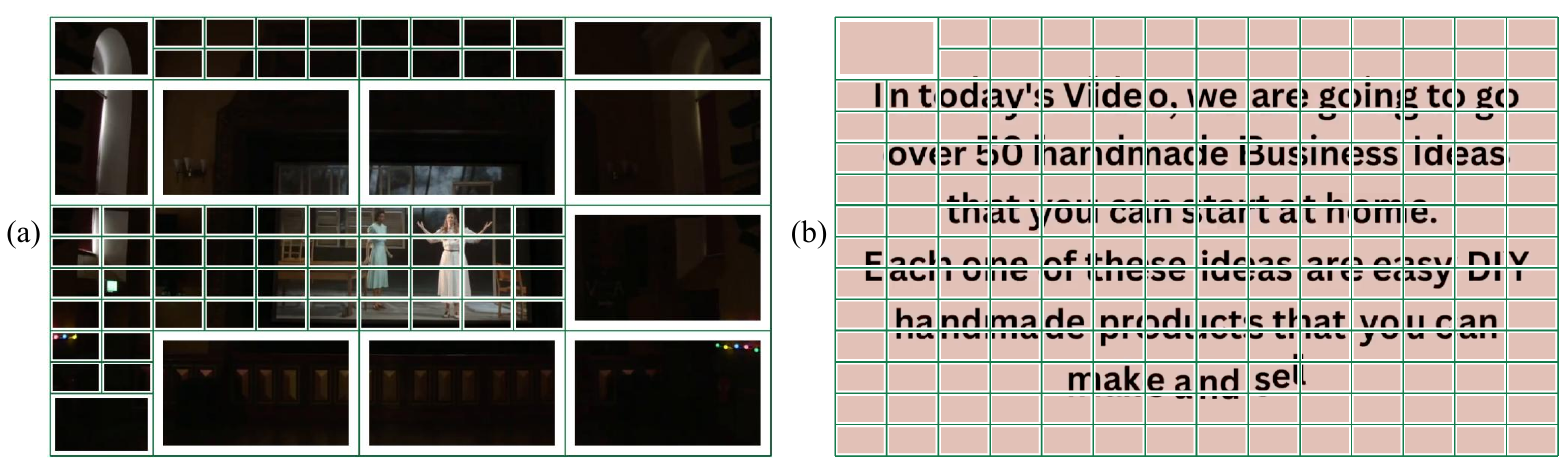}
\vspace{-3mm}

\caption{
Visualization of spatial token merging results.
Each image patch within a green box represents a single token.
}
\label{fig:visual_spatial}

\vspace{-3mm}
\end{figure*}

\begin{figure*}[t!]
\centering
\includegraphics[width=\linewidth]{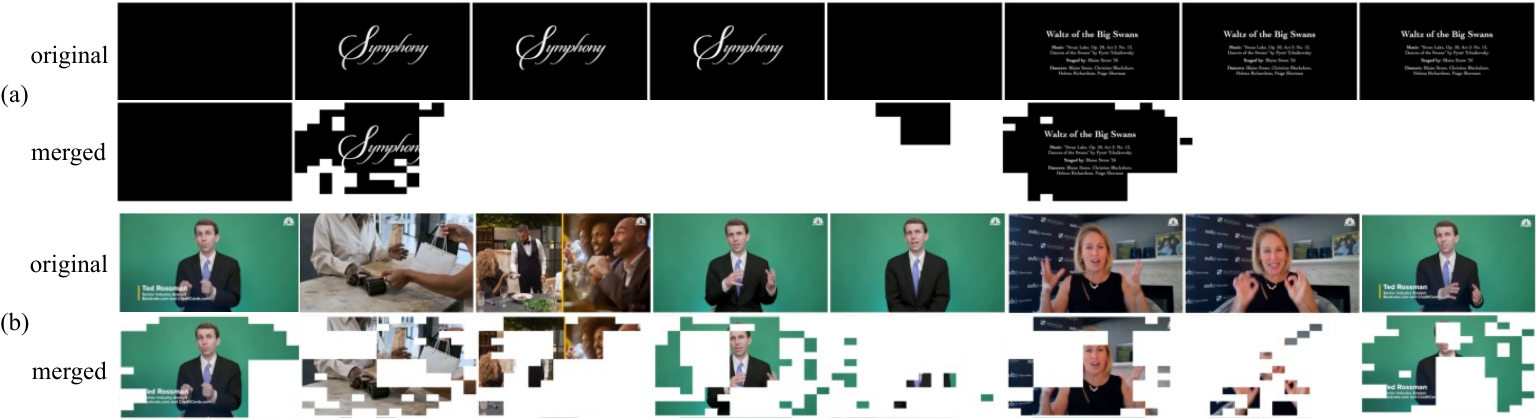}
\vspace{-5mm}

\caption{
Visualization of spatio-temporal token merging results on VideoMME.
(a) The first eight consecutive frames are sampled.
(b) Intermediate frames are sampled for illustration purpose.
Empty regions indicate areas that have been merged with early tokens.
}
\label{fig:visual_spatiotemporal}

\vspace{-5mm}
\end{figure*}

\noindent \textbf{Accuracy vs. Compression.}
\cref{fig:acc_token_tradeoff} illustrates the trade-off between accuracy and visual token retention ratio. 
On both datasets, our method consistently outperforms other query-agnostic token reduction methods~\cite{bolya2023tome, tao2025dycoke} across token retention levels, ranging from 50\% to 15\%.
This demonstrates the robustness of our method in achieving a favorable balance between compression efficiency and accuracy.

\subsection{Visualization of Token Merging}
Unlike uniform reduction methods, STTM adaptively reduces tokens based on the redundancy present in each video.
It allocates more tokens to regions that require fine-grained detail, while aggressively reducing tokens in redundant areas.
On VideoMME (30\% budget), its token retention ranges from 3.3\% to 51.2\% across videos, demonstrating its adaptability to varying content complexity.

\cref{fig:visual_spatial} presents the results of spatial merging before applying temporal merging.
\cref{fig:visual_spatial}~(a) illustrates a case where semantically similar tokens are effectively merged, reducing 196 visual tokens to 63 in a frame.
In contrast, \cref{fig:visual_spatiotemporal}~(b) shows a worst-case scenario for spatial compression.
In this frame, fine-grained tokens are preserved to retain small text details.
Although the compression rate is low, OCR accuracy on VideoMME is maintained after token merging.
While using all tokens yields an OCR accuracy of 67.6, ToMe, DyCoke-stage1, and our method achieve 58.3, 64.7, and 65.5, respectively, under token reduction.

\cref{fig:visual_spatiotemporal} shows the results of spatio-temporal merging.
In \cref{fig:visual_spatiotemporal}~(a), regions with newly emerging content remain intact, while redundant regions across consecutive frames are merged.
For instance, letter regions that appear in the second and sixth frames are preserved, while background tokens, such as the black areas, are merged into the earlier frames.
Duplicated frames (\eg, frames 3, 4, 7, and 8) are almost entirely merged.
\cref{fig:visual_spatiotemporal}~(b) illustrates an example with dynamic scene changes.
Similarly, within each scene, later frames tend to be merged into the first frame of the segment (\eg, frames 4 and 5).
Notably, subtle but semantically meaningful changes -- such as facial expressions or hand gestures -- are preserved in frames 6 and 7.

\section{Conclusion}
This is the first work to explore multi-granular video token merging in a training-free manner for video LLMs.
We propose a decomposed spatio-temporal token merging method, called STTM.
Its effectiveness is validated across six video QA benchmarks.
Notably, it largely outperforms others on VNBench which requires fine-grained understanding.
Since our method operates independently of user instructions, the KV cache can be reused across different questions for the same video, improving computational efficiency.
While our method successfully minimizes spatio-temporal redundancy, its performance currently depends on manually adjusted threshold values.
Exploring adaptive threshold selection is a promising direction, enabling automatic adjustment of token merging based on the given token budget.

\noindent\textbf{Acknowledgements.}
{\small
This work was partly supported by the NAVER Cloud Corporation.
}

{
    \small
    \bibliographystyle{ieeenat_fullname}
    \bibliography{main}
}

\clearpage

\appendix

\onecolumn

\section*{Appendix}
\cref{tab:supp_llava_video_7b,tab:supp_llava_onevision_7b,tab:supp_qwen2vl_7b,tab:supp_llava_video_72b} show the absolute values for the main comparison results.

\begin{table*}[bh!]
\aboverulesep=0ex
\belowrulesep=0ex
\setlength{\tabcolsep}{1pt} 
\centering
\resizebox{\linewidth}{!}{
\begin{tabular}{c | l | c | ccc | ccc | ccc | ccc | ccc | ccc | ccc}
\toprule
    \multirow{2}{*}{\begin{tabular}[c]{@{}l@{}} Token \\ Budget \end{tabular}} & \multirow{2}{*}{Method} & \multirow{2}{*}{\begin{tabular}[c]{@{}c@{}} Q. \\ Agn. \end{tabular}}
& \multicolumn{3}{c|}{\cellcolor{needleGreen} VNBench} & \multicolumn{3}{c|}{\cellcolor{longBlue} VideoMME} & \multicolumn{3}{c|}{\cellcolor{longBlue} LongVideoBench} & \multicolumn{3}{c|}{\cellcolor{longBlue} MLVU} &
\multicolumn{3}{c|}{\cellcolor{shortPink} EgoSchema} & \multicolumn{3}{c|}{\cellcolor{shortPink} NExT-QA} & \multicolumn{3}{c}{\cellcolor{avgColor} \textbf{\textit{Avg.}}}
\\
\cmidrule{4-24}
& & & Acc $\uparrow$ & TTFT $\downarrow$ & N$_\text{V}$ $\downarrow$ & Acc $\uparrow$ & TTFT $\downarrow$ & N$_\text{V}$ $\downarrow$ & Acc $\uparrow$ & TTFT $\downarrow$ & N$_\text{V}$ $\downarrow$ & Acc $\uparrow$ & TTFT $\downarrow$ & N$_\text{V}$ $\downarrow$ & Acc $\uparrow$ & TTFT $\downarrow$ & N$_\text{V}$ $\downarrow$ & Acc $\uparrow$ & TTFT $\downarrow$ & N$_\text{V}$ $\downarrow$ & Acc $\uparrow$ & TTFT $\downarrow$ & N$_\text{V}$ $\downarrow$ 
\\
\midrule
\midrule
\rowcolor{lightGray}
100\% & \textit{LLaVA-Video 7B} & \cmark &
    77.6 & 0.962 & 11149 & 63.1 & 2.039 & 22086 & 59.6 & 1.805 & 19624 & 70.9 & 2.343 & 25088 & 58.7 & 2.312 & 25069 & 82.9 & 0.659 & 8116 & 68.8 & 1.687 & 18522
\\
\midrule
\multirow{6}{*}{\begin{tabular}[c]{@{}l@{}} 50\% \end{tabular}}
& + FastV &  &
    72.7 & 0.503 & 5575 & 61.0 & 1.034 & 11043 & 57.4 & 0.918 & 9812 & 68.3 & 1.164 & 12544 & 57.6 & 1.166 & 12535 & 82.4 & 0.353 & 4058 & 66.6 & 0.856 & 9261
\\
& + DyCoke &  & 
    72.1 & 0.458 & 5386 & 61.5 & 0.912 & 10555 & 58.1 & 0.820 & 9393 & 69.5 & 1.046 & 11978 & 58.6 & 1.049 & 11969 & 82.1 & 0.330 & 3957 & 67.0 & 0.769 & 8873
\\
& + FrameFusion &  &
    76.2 & 0.471 & 5529 & 62.0 & 0.961 & 10739 & 59.2 & 0.853 & 9616 & 69.4 & 1.083 & 12138 & 57.7 & 1.088 & 12298 & 82.6 & 0.336 & 4032 & 67.9 & 0.799 & 9059
\\ \cmidrule{2-24}
& + ToMe & \cmark &
    74.1 & 0.518 & 5575 & 61.4 & 1.043 & 11043 & 58.0 & 0.949 & 9812 & 69.7 & 1.192 & 12544 & 58.7 & 1.199 & 12535 & 82.6 & 0.370 & 4058 & 67.4 & 0.878 & 9261
\\
& + DyCoke-stage1 & \cmark & 
    73.2 & 0.495 & 5386 & 62.0 & 0.981 & 10555 & 58.2 & 0.877 & 9393 & 69.7 & 1.123 & 11978 & 58.7 & 1.118 & 11969 & 82.4 & 0.350 & 3957 & 67.4 & 0.824 & 8873
\\
\rowcolor{lavendar}\cellcolor{white}
& + STTM (Ours) & \cmark & 
    77.4 & 0.455 & 4804 & 62.6 & 1.021 & 10771 & 59.6 & 0.895 & 9183 & 69.9 & 1.152 & 12187 & 58.6 & 1.045 & 10737 & 82.5 & 0.322 & 3452 & 68.4 & 0.815 & 8522
\\
\midrule
\multirow{6}{*}{\begin{tabular}[c]{@{}l@{}} 30\% \end{tabular}}
& + FastV &  & 
    61.6 & 0.336 & 3345 & 59.2 & 0.683 & 6626 & 54.7 & 0.610 & 5887 & 69.3 & 1.620 & 17562 & 56.9 & 0.771 & 7520 & 81.6 & 0.240 & 2435 & 63.9 & 0.710 & 7229
\\
& + DyCoke &  & 
    55.9 & 0.308 & 3548 & 60.7 & 0.598 & 6878 & 57.1 & 0.544 & 6131 & 67.3 & 0.684 & 7798 & 58.3 & 0.693 & 7792 & 81.5 & 0.229 & 2631 & 63.5 & 0.509 & 5797
\\
& + FrameFusion &  & 
    72.4 & 0.292 & 3157 & 60.7 & 0.581 & 6018 & 57.1 & 0.520 & 5462 & 67.5 & 0.650 & 6768 & 57.4 & 0.657 & 6870 & 81.9 & 0.218 & 2313 & 66.2 & 0.486 & 5098
\\ \cmidrule{2-24}
& + ToMe & \cmark &
    64.1 & 0.364 & 3345 & 59.2 & 0.720 & 6626 & 56.3 & 0.658 & 5888 & 67.0 & 0.821 & 7527 & 57.4 & 0.834 & 7521 & 81.6 & 0.265 & 2436 & 64.3 & 0.610 & 5557
\\
& + DyCoke-stage1 & \cmark & 
    63.9 & 0.358 & 3548 & 60.0 & 0.700 & 6878 & 56.5 & 0.631 & 6131 & 68.6 & 0.796 & 7798 & 58.7 & 0.800 & 7792 & 81.7 & 0.256 & 2631 & 64.9 & 0.590 & 5797
\\
\rowcolor{lavendar}\cellcolor{white}
& + STTM (Ours) & \cmark & 
    76.0 & 0.299 & 2649 & 62.3 & 0.640 & 5929 & 57.0 & 0.616 & 5702 & 68.5 & 0.769 & 7337 & 58.0 & 0.773 & 7285 & 82.0 & 0.235 & 2168 & 67.3 & 0.555 & 5179
\\
\bottomrule
\end{tabular}
}
\vspace{-3mm}
\caption{
Comparison of training-free token reduction methods using LLaVA-Video-7B under 50\% and 30\% pre-filling token budgets.
}
\label{tab:supp_llava_video_7b}
\vspace{-3mm}
\end{table*}

\begin{table*}[bh!]
\aboverulesep=0ex
\belowrulesep=0ex
\setlength{\tabcolsep}{1pt} 
\centering
\resizebox{\linewidth}{!}{
\begin{tabular}{c | l | c | ccc | ccc | ccc | ccc | ccc | ccc | ccc}
\toprule
    \multirow{2}{*}{\begin{tabular}[c]{@{}l@{}} Token \\ Budget \end{tabular}} & \multirow{2}{*}{Method} & \multirow{2}{*}{\begin{tabular}[c]{@{}c@{}} Q. \\ Agn. \end{tabular}}
& \multicolumn{3}{c|}{\cellcolor{needleGreen} VNBench} & \multicolumn{3}{c|}{\cellcolor{longBlue} VideoMME} & \multicolumn{3}{c|}{\cellcolor{longBlue} LongVideoBench} & \multicolumn{3}{c|}{\cellcolor{longBlue} MLVU} &
\multicolumn{3}{c|}{\cellcolor{shortPink} EgoSchema} & \multicolumn{3}{c|}{\cellcolor{shortPink} NExT-QA} & \multicolumn{3}{c}{\cellcolor{avgColor} \textbf{\textit{Avg.}}}
\\
\cmidrule{4-24}
& & & Acc $\uparrow$ & TTFT $\downarrow$ & N$_\text{V}$ $\downarrow$ & Acc $\uparrow$ & TTFT $\downarrow$ & N$_\text{V}$ $\downarrow$ & Acc $\uparrow$ & TTFT $\downarrow$ & N$_\text{V}$ $\downarrow$ & Acc $\uparrow$ & TTFT $\downarrow$ & N$_\text{V}$ $\downarrow$ & Acc $\uparrow$ & TTFT $\downarrow$ & N$_\text{V}$ $\downarrow$ & Acc $\uparrow$ & TTFT $\downarrow$ & N$_\text{V}$ $\downarrow$ & Acc $\uparrow$ & TTFT $\downarrow$ & N$_\text{V}$ $\downarrow$
\\
\midrule
\midrule
\rowcolor{lightGray}
100\% & \textit{LLaVA-OV 7B} & \cmark &
68.8 & 0.922 & 11149 & 
59.0 & 1.904 & 22086 & 
56.3 & 1.712 & 19624 &
67.5 & 2.224 & 25088 &
61.3 & 2.216 & 25069 &
80.6 & 0.630 & 8116 &
65.6 & 1.601 & 18522
\\
\midrule
\multirow{6}{*}{\begin{tabular}[c]{@{}l@{}} 50\% \end{tabular}}
& + FastV &  &
    65.6 & 0.467 & 5575 & 58.4 & 0.934 & 11043 & 56.2 & 0.842 & 9812 & 67.4 & 1.071 & 12544 & 61.0 & 1.080 & 12535 & 80.2 & 0.330 & 4058 & 64.8 & 0.787 & 9261
\\
& + DyCoke &  & 
    64.9 & 0.418 & 5386 & 60.1 & 0.831 & 10555 & 57.9 & 0.740 & 9393 & 68.7 & 0.941 & 11978 & 61.8 & 0.948 & 11969 & 80.4 & 0.301 & 3957 & 65.6 & 0.697 & 8873
\\
& + FrameFusion &  &
    67.8 & 0.436 & 5568 & 59.8 & 0.865 & 10888 & 56.6 & 0.778 & 9719 & 68.2 & 0.994 & 12379 & 60.8 & 0.997 & 12511 & 80.8 & 0.312 & 4056 & 65.7 & 0.730 & 9187
\\ \cmidrule{2-24}
& + ToMe & \cmark &
    67.1 & 0.483 & 5575 & 59.8 & 0.963 & 11043 & 57.7 & 0.865 & 9812 & 68.8 & 1.101 & 12544 & 62.3 & 1.108 & 12535 & 80.1 & 0.343 & 4058 & 66.0 & 0.811 & 9261
\\
& + DyCoke-stage1 & \cmark & 
    65.1 & 0.459 & 5386 & 59.3 & 0.911 & 10555 & 58.2 & 0.821 & 9393 & 69.1 & 1.039 & 11978 & 61.8 & 1.049 & 11969 & 80.6 & 0.327 & 3957 & 65.7 & 0.767 & 8873
\\
\rowcolor{lavendar}\cellcolor{white}
& + STTM (Ours) & \cmark & 
    71.2 & 0.404 & 4573 & 60.7 & 0.773 & 8579 & 57.7 & 0.804 & 9011 & 69.7 & 1.020 & 11692 & 61.7 & 0.972 & 10944 & 80.5 & 0.312 & 3628 & 66.9 & 0.714 & 8071
\\
\midrule
\multirow{6}{*}{\begin{tabular}[c]{@{}l@{}} 30\% \end{tabular}}
& + FastV &  & 
    59.2 & 0.303 & 3345 & 58.1 & 0.597 & 6626 & 55.3 & 0.543 & 5887 & 65.6 & 0.681 & 7526 & 60.9 & 0.694 & 7520 & 79.6 & 0.218 & 2435 & 63.1 & 0.506 & 5556
\\
& + DyCoke &  & 
    53.9 & 0.280 & 3548 & 59.8 & 0.537 & 6878 & 55.9 & 0.488 & 6131 & 68.3 & 0.598 & 7798 & 62.1 & 0.607 & 7792 & 79.7 & 0.208 & 2631 & 63.3 & 0.453 & 5797
\\
& + FrameFusion &  & 
    66.1 & 0.261 & 3263 & 59.0 & 0.499 & 6154 & 56.5 & 0.455 & 5566 & 66.7 & 0.570 & 6914 & 60.3 & 0.582 & 7251 & 79.9 & 0.195 & 2400 & 64.7 & 0.427 & 5258
\\ \cmidrule{2-24}
& + ToMe & \cmark &
    59.1 & 0.332 & 3345 & 59.8 & 0.649 & 6626 & 56.5 & 0.587 & 5888 & 69.1 & 0.740 & 7527 & 62.1 & 0.750 & 7521 & 79.5 & 0.240 & 2436 & 64.4 & 0.550 & 5557
\\
& + DyCoke-stage1 & \cmark & 
    58.4 & 0.324 & 3548 & 60.3 & 0.633 & 6878 & 56.2 & 0.577 & 6131 & 68.0 & 0.723 & 7798 & 62.3 & 0.731 & 7792 & 79.7 & 0.236 & 2631 & 64.1 & 0.537 & 5797
\\
\rowcolor{lavendar}\cellcolor{white}
& + STTM (Ours) & \cmark & 
    70.4 & 0.277 & 2773 & 60.6 & 0.601 & 6264 & 56.6 & 0.570 & 6022 & 68.4 & 0.735 & 7989 & 61.8 & 0.570 & 5621 & 79.7 & 0.240 & 2577 & 66.2 & 0.499 & 5208
\\
\bottomrule
\end{tabular}
}
\vspace{-3mm}
\caption{
Comparison of training-free token reduction methods using LLaVA-OneVision-7B.
}
\label{tab:supp_llava_onevision_7b}
\vspace{-1mm}
\end{table*}

\begin{table*}[tbh!]
\aboverulesep=0ex
\belowrulesep=0ex
\setlength{\tabcolsep}{1pt} 
\centering
\begin{minipage}{0.70\textwidth}
\resizebox{\linewidth}{!}{
    \begin{tabular}{c | l | ccc | ccc | ccc | ccc }
    \toprule
        \multirow{2}{*}{\begin{tabular}[c]{@{}l@{}} Token \\ Budget \end{tabular}} & \multirow{2}{*}{Method} & \multicolumn{3}{c|}{\cellcolor{needleGreen} VNBench} & \multicolumn{3}{c|}{\cellcolor{longBlue} VideoMME} & \multicolumn{3}{c|}{\cellcolor{longBlue} LongVideoBench} & \multicolumn{3}{c}{\cellcolor{avgColor} \textbf{\textit{Avg.}}}
    \\
    \cmidrule{3-14}
    
    & & Acc $\uparrow$ & TTFT $\downarrow$ & N$_\text{V}$ $\downarrow$ & Acc $\uparrow$ & TTFT $\downarrow$ & N$_\text{V}$ $\downarrow$ & Acc $\uparrow$ & TTFT $\downarrow$ & N$_\text{V}$ $\downarrow$ & Acc $\uparrow$ & TTFT $\downarrow$ & N$_\text{V}$ $\downarrow$ 
    \\
    \midrule
    \midrule
    \rowcolor{lightGray}
    100\% 
     &  \textit{Qwen2VL 7B} &
        66.4 & 2.438 & 22025 & 61.8 & 10.745 & 74982 & 56.8 & 10.597 & 72109 & 61.7 & 7.927 & 56372
    \\
    \midrule
    \multirow{3}{*}{\begin{tabular}[c]{@{}l@{}} 50\% \end{tabular}}
    & + ToMe &
        63.4 & 1.130 & 11013 & 61.9 & 4.509 & 37491 & 56.7 & 4.348 & 36054 & 60.7 & 3.329 & 28186
    \\
    & + DyCoke-stage1 &
        65.1 & 1.049 & 10645 & 62.6 & 4.166 & 36057 & 57.5 & 4.148 & 34735 & 61.7 & 3.121 & 27146
    \\
    \rowcolor{lavendar}\cellcolor{white}
    & + STTM (Ours) & 
        69.8 & 0.726 & 7228 & 62.9 & 4.761 & 39217 & 57.4 & 4.610 & 37315 & 63.4 & 3.366 & 27920
    \\
    \midrule
    \multirow{3}{*}{\begin{tabular}[c]{@{}l@{}} 30\% \end{tabular}}
    & + ToMe &
        56.9 & 0.740 & 6608 & 61.4 & 2.835 & 22496 & 54.7 & 2.600 & 21633 & 57.6 & 2.058 & 16912
    \\
    & + DyCoke-stage1 & 
        54.2 & 0.706 & 6980 & 61.8 & 2.710 & 23479 & 57.5 & 2.694 & 22649 & 57.8 & 2.037 & 17703
    \\
    \rowcolor{lavendar}\cellcolor{white}
    & + STTM (Ours) & 
        66.7 & 0.420 & 6748 & 62.4 & 2.766 & 23143 & 56.9 & 2.472 & 20022 & 62.0 & 1.886 & 16638
    \\
    \bottomrule
    \end{tabular}
}
\vspace{-3mm}
\caption{Comparison using Qwen2VL-7B.
Relative to \colorbox{lightGray}{100\% result}.
}
\label{tab:supp_qwen2vl_7b}
\vspace{-5mm}
\end{minipage}
\hfill
\begin{minipage}{0.28\linewidth}
\resizebox{1.0\linewidth}{!}{
    \begin{tabular}{c | l | ccc }
    \toprule
        \multirow{2}{*}{\begin{tabular}[c]{@{}l@{}} Token \\ Budget \end{tabular}} & \multirow{2}{*}{Method}
    & \multicolumn{3}{c}{\cellcolor{longBlue} VideoMME}
    \\
    \cmidrule{3-5}
    & & Acc $\uparrow$ & TTFT $\downarrow$ & N$_\text{V}$ 
    \\
    \midrule
    \midrule
    \rowcolor{lightGray}
    100\% & \textit{LLaVA-Video 72B} &
        70.5 & 17.698 & 22086
    \\
    \midrule
    \multirow{3}{*}{\begin{tabular}[c]{@{}l@{}} 50\% \end{tabular}}
    & + ToMe &
        70.6 & 8.424 & 11043
    \\
    & + DyCoke-stage1 & 
        70.4 & 8.052 & 10555
    \\
    \rowcolor{lavendar}\cellcolor{white}
    & + STTM (Ours) &
        71.4 & 7.821 & 10082
    \\
    \midrule
    \multirow{3}{*}{\begin{tabular}[c]{@{}l@{}} 30\% \end{tabular}}
    & + ToMe & 
        68.5 & 5.186 & 6626
    \\
    & + DyCoke-stage1 & 
        69.3 & 5.353 & 6878
    \\
    \rowcolor{lavendar}\cellcolor{white}
    & + STTM (Ours) & 
        69.9 & 5.405 & 6897
    \\
    \bottomrule
    \end{tabular}
}
\vspace{-3mm}
\caption{
Comparison using LLaVA-Video-72B.
Relative to \colorbox{lightGray}{100\% result}.
}
\label{tab:supp_llava_video_72b}
\vspace{-5mm}
\end{minipage}
\end{table*}

\end{document}